\documentclass{article}

\usepackage{arxiv}

\usepackage[utf8]{inputenc} 
\usepackage[T1]{fontenc}    
\usepackage{hyperref}       
\usepackage{url}            
\usepackage{booktabs}       
\usepackage{amsfonts}       
\usepackage{nicefrac}       
\usepackage{microtype}      
\usepackage{xcolor} 
\usepackage{cite}
\usepackage{graphicx}
\usepackage{float} 
\usepackage{enumitem}
\usepackage{amsmath}
\usepackage{booktabs} 
\usepackage{multirow} 
\usepackage{amsmath} 
\usepackage{pifont}
\usepackage{wrapfig}

\title{HCRMP: A LLM-Hinted Contextual Reinforcement Learning Framework for Autonomous Driving}

\author{
  Zhiwen Chen\textsuperscript{1} \quad Bo Leng\textsuperscript{1} \quad Zhuoren Li\textsuperscript{1} \quad Hanming Deng\textsuperscript{2} 
  \AND
  \quad Guizhe Jin\textsuperscript{1} \quad Ran Yu\textsuperscript{1} \quad Huanxi Wen\textsuperscript{1} \\[1.5ex]
  Tongji University\textsuperscript{1}, SenseTime Research\textsuperscript{2} \\[1.5ex]
  {\{2411451, lengbo, 1911055, ranyu, 2311385\}@tongji.edu.cn} \\
  {jgz13573016892@163.com} \quad {denghanming@sensetime.com}
}

\begin{document}

\maketitle
 
\begin{abstract}


Integrating the understanding and reasoning capabilities of Large Language Models (LLM) with the self-learning capabilities of Reinforcement Learning (RL) enables more reliable driving performance under complex driving conditions.
There has been a lot of work exploring LLM-Dominated RL methods in the field of autonomous driving motion planning.
These methods, which utilize LLM to directly generate policies or provide decisive instructions during policy learning of RL agent, are centrally characterized by an over-reliance on LLM outputs.
However, LLM outputs are susceptible to hallucinations.
Evaluations show that state-of-the-art LLM indicates a non-hallucination rate of only approximately 57.95\% when assessed on essential driving-related tasks.
Thus, in these methods, hallucinations from the LLM can directly jeopardize the performance of driving policies.
This paper argues that \textit{maintaining relative independence between the LLM and the RL} is vital for solving the hallucinations problem.
Consequently, this paper is devoted to propose a novel \textbf{LLM-Hinted RL} paradigm.
The LLM is used to generate semantic hints for state augmentation and policy optimization to assist RL agent in motion planning, while the RL agent counteracts potential erroneous semantic indications through policy learning to achieve excellent driving performance.
Based on this paradigm, we propose the \textbf{HCRMP} (LLM-Hinted Contextual Reinforcement Learning Motion Planner) architecture, which is designed that includes \ding{172} 
Augmented Semantic Representation Module to extend state space.
\ding{173} Contextual Stability Anchor Module enhances the reliability of multi-critic weight hints by utilizing information from the knowledge base.
\ding{174} Semantic Cache Module is employed to seamlessly integrate LLM low-frequency guidance with RL high-frequency control.
Extensive experiments in CARLA validate HCRMP's strong overall driving performance.
HCRMP achieves a task success rate of up to 80.3\% under diverse driving conditions with different traffic densities. 
Under safety-critical driving conditions, HCRMP significantly reduces the collision rate by 11.4\%, which effectively improves the driving performance in complex scenarios.
\end{abstract}

\section{Introduction}



Reinforcement learning (RL) is the method for learning optimal policies by maximizing expected returns through interactions with the environment. It has proven to be effective for solving complex decision-making problems and has attracted significant attention in various fields~\cite{shakya2023reinforcement,wang2022deep,gu2022review}.
For the motion planning task in autonomous driving (AD), the RL agent can dynamically generate trajectories or control commands that follow the learned driving policy, based on the fused multimodal traffic features\cite{chen2022milestones,dang2023event,alighanbari2022deep}. 
However, RL has limited understanding and reasoning in complex driving conditions and often fails to identify critical traffic features\cite{brown2020language,touvron2023llama}, which can result in unreliable driving actions. 
In contrast, large language models (LLM) poss strong semantic understanding and common-sense reasoning abilities\cite{rajani2019explain,huang2023can}, and have achieved significant advances in multitasking in recent years\cite{grattafiori2024llama,bi2024deepseek,guo2025deepseek}. 
AD systems require the extensive common sense and high-level decision-making capabilities, which are exactly what large language models can provide\cite{fu2024limsim++,ma2024lampilot,yang2024drivearena,chen2024driving,yang2023llm4drive}. 
Therefore, integrating LLM and RL within a unified AD system leverages the strengths of both approaches\cite{cao2024survey}. This integration enhances policy understanding and reasoning during self-learning\cite{schoepp2025evolving}, leading to more reliable and safer driving actions in complex driving conditions\cite{wu2024robust}.



Despite the significant potential benefits of integrating LLM and RL for AD, effective integration remains a critical and unresolved challenge. Current methods primarily use RL agent to assist in the policy optimization of LLM \cite{yildirim2024highwayllm,jiang2025alphadrive,sun2024optimizing} or employ LLM to directly instruct RL agent policy generation \cite{huang2024vlm,sheng2025curricuvlm,pang2024large,peng2025learningflow,han2024autoreward,doroudian2024clip,ye2025lord,hazra2024revolve}.
In the former approach, the LLM outputs directly generate the driving policy, while in the latter, the LLM strongly provides decisive instructions during the policy learning process of the RL agent.
Because both methods demonstrate strong reliance on the LLM, we refer to such methods as LLM-Dominated RL Methods.

\begin{figure}[H] 
    \centering
    \includegraphics[width=1.0\linewidth]{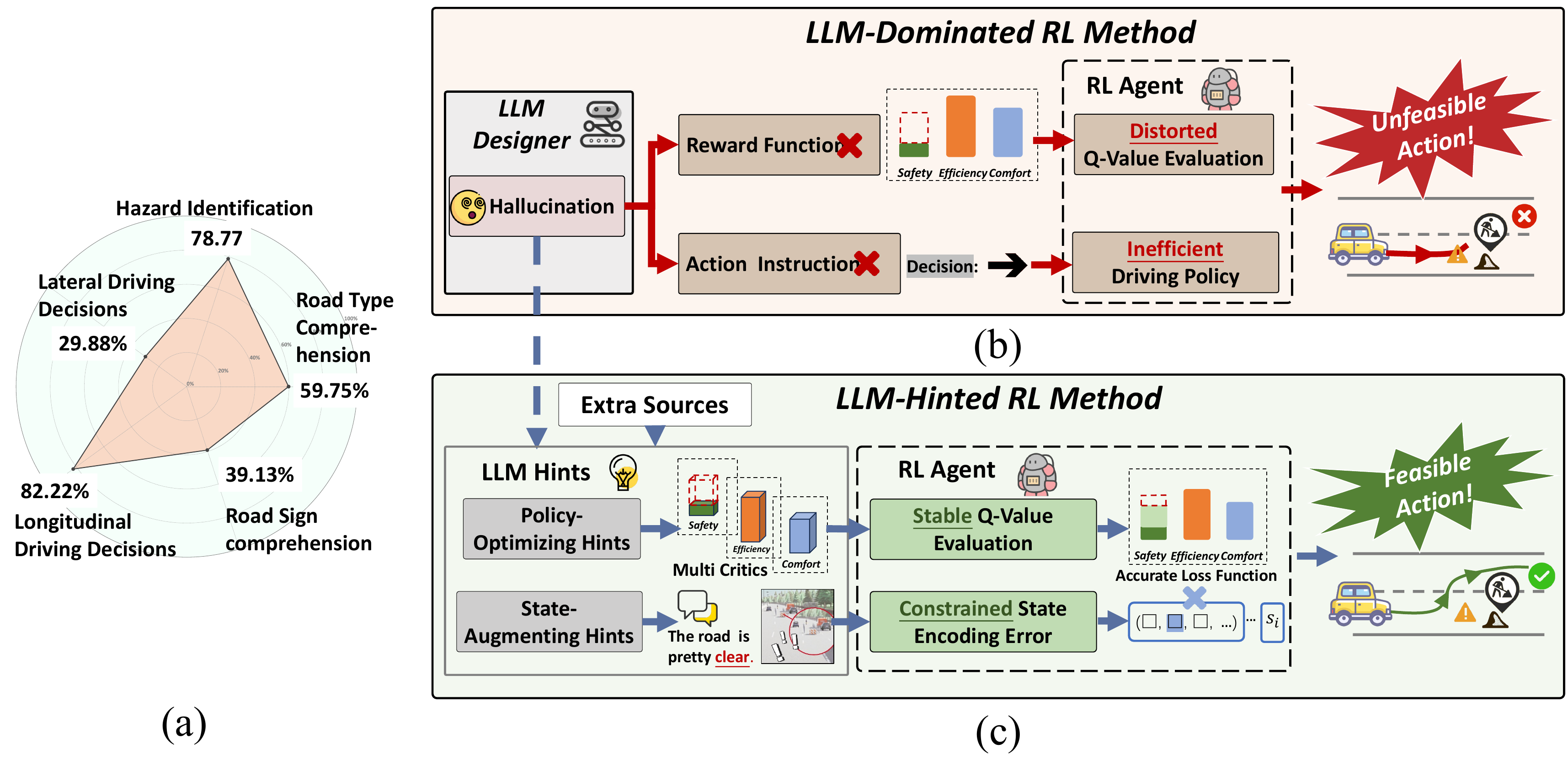}
    \caption{LLM performance evaluation and hallucination impact on LLM-RL methods.
    Figure (a) shows the SOTA LLM's non-hallucination rates across five driving tasks.
    Figure (b) illustrates the LLM-Dominated RL methods. LLM hallucinations directly distort the RL agent's Q-value estimation and degrade policy efficiency, leading to unfeasible driving actions.
    Figure (c) presents the LLM-Hinted RL method. LLM provides semantic hints to the RL agent instead of directly dictating decisions. The RL agent, through its own policy learning process, effectively buffers the negative impact of these hallucinations, thereby preventing unfeasible actions.}
    \label{fig:intro_1}
\end{figure}

However, LLM outputs are known to be susceptible to hallucinations \cite{orgad2024llms,sun2024crosscheckgpt,zhang2024llm,liu2024exploring,sriramanan2024llm}, which can distort decision-making and compromise policy stability.
The Gemini-2.5-Pro model, a state-of-the-art (SOTA) LLM \cite{li2025sti}, is evaluated on five key dimensions related to two essential driving capabilities: scenario understanding and action response, as illustrated in Figure~\ref{fig:intro_1} (a). 
The result shows a non-hallucination rate of only 57.95\%, implying that over 40\% of its outputs are prone to hallucinations.
As shown in Figure~\ref{fig:intro_1} (b), when LLM hallucinations occur, these incorrect signals are directly propagated to the downstream decision-making process.
This distorts the Q-value evaluation and compromises the safety and stability of the policy.    
The direct result is a significant degradation in overall driving performance. It is thus imperative for AD systems combining LLM and RL to mitigate the policy instability caused by LLM hallucinations.

This paper argues that \textit{maintaining relative independence between the LLM and the RL} is an essential way to solve the hallucinations problem.
The fundamental reason is that such separation preserves the RL agent’s autonomy in decision-making and adaptation, while allowing the LLM to provide semantic hints as auxiliary inputs and intrinsic modulation. 
Accordingly, We propose an LLM-Hinted RL motion planning paradigm, as illustrated in Figure~\ref{fig:intro_1} (c). 
Even if the LLM outputs are unstable, the RL agent is able to counteract potential erroneous semantic indications through policy learning, avoiding the direct generation of unreasonable actions.
At the same time, this separated structure preserves the utilization of LLM strengths in driving conditions comprehension and common-sense reasoning, enabling context-aware guidance for the RL agent in a way that maintains its fundamental self-optimization capabilities. 

Based on the LLM-Hinted RL paradigm, we develop an architecture called HCRMP (LLM-\textbf{H}inted \textbf{C}ontextual \textbf{R}einforcement Learning \textbf{M}otion \textbf{P}lanner). The architecture comprises three key components: \ding{172} the Augmented Semantic Representation Module, which utilizes semantic hints from the LLM to extend the state space; \ding{173} the Contextual Stability Anchor Module, which leverages information from the structured knowledge base to improve the reliability of the weight hints that the LLM generates for each critic network; and \ding{174} the Semantic Cache Module, which enables efficient and stable training through fixed-frequency hierarchical outputs and the historical context cache matching strategy.
The LLM provides state-augmenting and policy-optimizing semantic hints as auxiliary inputs and intrinsic modulation to the RL agent, rather than directly controlling policy generation. Meanwhile, the RL agent autonomously executes motion planning, and the LLM and RL modules collaborate asynchronously at different temporal scales, ensuring training stability.
In addition, extensive experiments in the CARLA simulator validate the effectiveness of our proposed HCRMP framework, highlighting its superior overall performance, particularly in demanding driving conditions.
HCRMP achieves a task success rate of up to \textbf{80.3\%} across diverse conditions with varied traffic densities. 
Critically, in safety-critical driving conditions, HCRMP achieves a significant \textbf{11.4\%}reduction in the collision rate.
The contributions of this study can be summarized as follows:
\begin{itemize}[leftmargin=1.5em]
    \item We classify existing LLM-Dominated RL methods, clarify their strong reliance on LLM outputs, and highlight the problem that hallucinations from the LLM can degrade driving performance.
    To address these challenges, we propose the LLM-Hinted RL paradigm.
    \item We propose a novel motion planning architecture named HCRMP. 
    By combining the semantic hints for state augmentation and
    policy optimization provided by LLM with the self-learning capabilities of RL, it significantly improves driving performance in diverse driving conditions.
    \item 
    Extensive experiments in CARLA validate HCRMP's strong overall driving performance.
    HCRMP achieves a task success rate of up to \textbf{80.3\%} under diverse driving conditions and, critically, reduces the collision rate by \textbf{11.4\%} in safety-critical driving conditions.
\end{itemize}
\section{Related Works} 
LLM-Dominated RL methods for AD motion planning fall into two main categories: RL-assisted LLM policy optimization and LLM-instructed RL policy generation. 
\subsection{RL-Assisted LLM Policy Optimization} 
This type of methods typically convert conditions information into linguistic inputs to generate action instructions or probabilities distributions \cite{zhang2023rladapter,li2022pre,shi2023unleashing,yan2023ask,yan2024efficient,carta2023grounding}. RL agent enables the fine-tuning of LLM parameters using reward signals from the environment, optimizing its policy to maximize cumulative rewards \cite{zhai2024fine,havrilla2024teaching}.
Recently, the integrated paradigm of LLM and RL has demonstrated considerable potential for motion planning in AD.
Existing studies primarily use LLM to generate trajectories or control commands, with RL submodules integrated for action suggestion generation or policy optimization. HighwayLLM \cite{yildirim2024highwayllm} drives an LLM agent using meta-actions output by a pre-trained RL model, combining the current state and similar trajectories to generate specific actions. AlphaDrive \cite{jiang2025alphadrive}, in contrast, leverages GRPO-based RL reward function to enhance the driving policy of its vision-language models (VLM).
This enhancement enables the VLM to better adapt to dynamic driving conditions.
While these methods can leverage the LLM strengths in complex scenario understanding and decision-making, their core risk is that any erroneous instructions generated by LLM can be directly mapped to unreasonable driving actions, which fundamentally threaten the driving safety.
\subsection{LLM-Instructed RL Policy Generation}  

For the other type of method, RL agent is used to generate control commands. LLM serves as a sub-module to support the policy optimization. Specifically, LLM is utilized to provide intrinsic rewards, which is proven to improve the learning efficiency of RL \cite{qu2025latent,xie2023text2reward,choi2022lmpriors,du2023guiding,ma2023eureka}.
For AD motion planning, LearningFlow \cite{peng2025learningflow} and Autoreward \cite{han2024autoreward} utilize a closed-loop framework in which the LLM automatically generates reward signals to guide RL agent training. Similarly, Clip-RLdrive \cite{doroudian2024clip}, LORD \cite{ye2025lord}, and REvolve \cite{hazra2024revolve} directly rely on the LLM to generate reward values for policy learning. 
However, these methods are highly sensitive to the quality of LLM outputs and are therefore quite susceptible to the adverse effects of hallucinations.
In contrast, our proposed method utilizes a \textit{weakly coupled} integration of the LLM and RL. This significantly diminishes the system's vulnerability to fluctuations in the LLM outputs, preserving the LLM inherent semantic and reasoning advantages while ensuring RL maintains its fundamental self-optimization capabilities.
\section{Methodology}
\begin{figure}[ht] 
    \centering
    \includegraphics[width=0.85\linewidth]{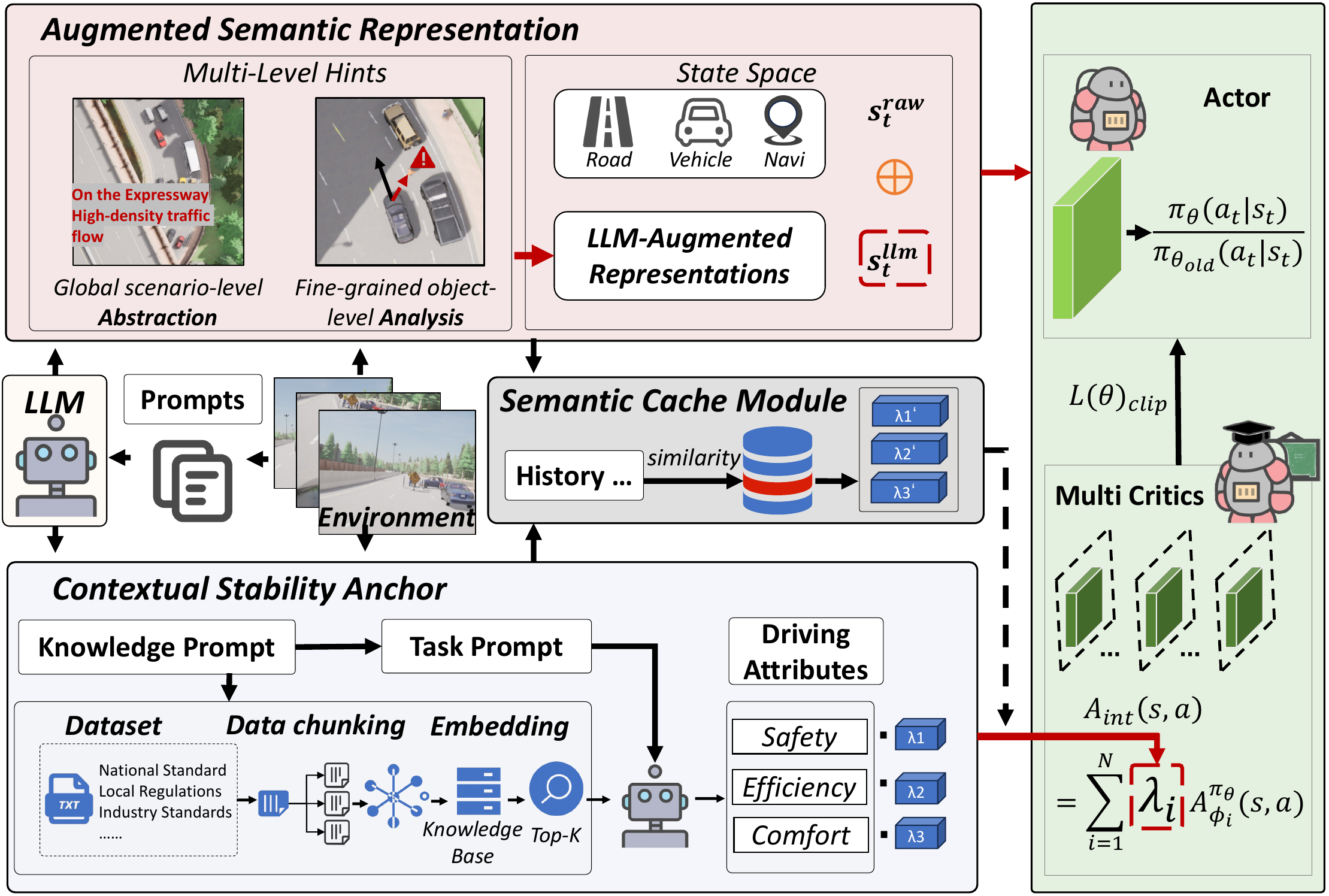}
    \caption{The framework of our proposed HCRMP. LLM acts in the Augmented Semantic Representation module to fetch information at the scenario level and object level, extending the state space. Meanwhile, LLM acts in the Contextual Stability Anchor module to generate reliable weights between multi critics, utilizing the knowledge base to mitigate the output fluctuations. When LLM fails to provide timely guidance, the Semantic Cache module replaces the missing weights by retrieving the most similar historical driving conditions.
    Hints from LLM are ultimately input to the RL agent's actor and multi-critic networks for optimal policy learning.}
    \label{fig:tech-overview}
\end{figure}

This section outlines the proposed HCRMP framework.
As illustrated in Figure ~\ref{fig:tech-overview}, HCRMP comprises three key components: the Augmented Semantic Representation Module (Section 3.2), the Contextual Stability Anchor (Section 3.3), and the Semantic Cache Module (Section 3.4).

The Augmented Semantic Representation (ASR) module employs LLM for hierarchical scenario reasoning, from global abstraction to object-level analysis. The resulting multi-level semantic hints are encoded into compact vectors and integrated into the RL agent’s state space to enhance driving condition awareness.
The Contextual Stability Anchor (CSA) Module utilizes external knowledge sources, including traffic regulations and fundamental priors, to generate more reliable semantic hints.
These hints then dynamically influence Q-value evaluation by modulating the impact of various driving attributes, thereby guiding the RL agent toward more effective decision-making.
Furthermore, the Semantic Cache Module(SCM) addresses the temporal mismatch between the low-frequency semantic guidance from the LLM and the high-frequency control commands of the RL.

\subsection{Problem Formulation}

We formulate the AD task as a Markov Decision Process (MDP), represented by the tuple $(S, A, P, R, \gamma)$. Here, the action space $A$ consists of two continuous control variables: throttle/brake control commands and steering angle, each constrained within the normalized range $[-1, +1]$. The state space $S$ is composed of two components: the raw state characteristics $s_t^{\text{raw}}$ and the LLM-augmented semantic state characteristics $s_t^{\text{llm}}$. The transition function $P(s' \mid s, a)$ defines the probability distribution over next states, the reward function $R$ provides feedback signals reflecting the quality of actions, and $\gamma \in [0,1]$ is a discount factor that trades off between immediate and future rewards.
The agent employs a deep neural network (DNN) policy, denoted by $\pi_\theta$, where $\theta$ represents the learnable parameters. At each time step $t$, given the current state $s_t \in S$, the policy samples a bounded continuous action $a_t \sim \pi_\theta(\cdot \mid s_t)$ from a Beta distribution, ensuring constrained control commands. Then, the agent executes action $a_t \in A$, receives an immediate reward $r_t \in R$, and transitions to the next state $s_{t+1}$ based on $P$.
The agent’s objective is to discover an optimal policy $\pi^\ast$ through interaction with the environment that maximizes the expected discounted cumulative return in \textit{Eq }\ref{policy}.
\begin{equation}
\pi^* = \arg\max_{\pi}\mathbb{E}\left[\sum_{t=0}^{\infty} \gamma^t r(s_t,a_t)\right]
\label{policy}
\end{equation}

\subsection{Augmented Semantic Representation}
The LLM-based ASR module is designed to provide the RL agent with multi-level semantic encodings, which in turn augments its state space.
The LLM performs hierarchical situational reasoning, transitioning from a global scenario-level abstraction to a fine-grained object-level analysis. 
Based on the scenario's topological configuration and traffic dynamics, the LLM executes semantic parsing to categorize road situations.
Then, this categorized understanding is encoded into a 4-dimensional scenario-level vector.
Concurrently, by analyzing the semantic characteristics of surrounding traffic participants, the LLM pinpoints critical agents.
Following this identification, it quantifies their relative spatial configurations and maps them to discrete directional categories.
These are then encoded as a 9-dimensional object-level vector. 
To ensure consistent representation, particularly in low-density traffic conditions, a semantic compensation mechanism can be implemented to preserve integrity.

At each time step $t$, the state space of the agent $S_t$ is constructed by combing heterogeneous feature sets, including raw sensory inputs $s_t^{\text{raw}}$ and LLM-enhanced semantic embeddings $s_t^{\text{llm}}$ 
- where $s_t^{\text{raw}} = (f_t^1, f_t^2, \dots, f_t^J)$, and each feature frame $f_t^j = (I_t^j, T_t^j, E_t^j, \mathcal{N}_t^j)$ comprises multi-view RGB image features $I_t^j$, road topology features $T_t^j$, ego-vehicle dynamics $E_t^j$, and navigation path embeddings $\mathcal{N}_t^j$.
The RL agent processes incoming multimodal inputs via a lightweight yet effective visual backbone (ShuffleNetV2 \cite{ma2018shufflenet}) for spatial encoding and Gated Recurrent Units (GRU) for temporal semantic modeling. In this pipeline, the semantic representation $s_t^{\text{llm}}$ functions as an auxiliary information source, serving as semantic guidance to enhance situational awareness.

\subsection{Contextual Stability Anchor}
To satisfy the demands for adaptive trade-offs among driving attributes (e.g., safety, comfort, efficiency), we design a multi-critic framework based on Proximal Policy Optimization (PPO) \cite{schulman2017proximal}. The framework decouples the Q-value evaluation of multiple driving attributes and enable independent representation learning for distinct objectives.
By formulating prompts, the LLM performs contextual analysis of dynamic traffic conditions and dynamically generates a set of adaptive weights $ \{ \lambda_i \}_{i=1}^N$, where $N$ is the number of attributes, $\lambda_i \in [0, 1]$, and $\sum_{i=1}^N \lambda_i = 1$, which are then used as part of an integrated advantage function $\hat{A}_{\text{int}}$ estimated via Generalized Advantage Estimation (GAE) in \textit{Eq }\ref{Advantage}
\begin{equation}
\hat{A}_{\text{int}}(s, a) = \sum_{i=1}^N \lambda_i A_i^{\pi_{\theta}}(s, a)\label{Advantage}
\end{equation}
The clipping objective is defined in \textit{Eq }\ref{L_clip}, where the $\text{clip}(\cdot)$ function constrains the update range to stabilize training.
$ratio_t(\theta)$ denotes the probability ratio between the current policy $\pi_{\theta}$ and the previous policy $\pi_{\theta_{\text{old}}}$.
\begin{equation}
L_{\text{clip}}(\theta) = \mathbb{E}_t \left[ \min \left( ratio_t(\theta) \hat{A}_{\text{int}}, \, \text{clip}(ratio_t(\theta), 1 - \varepsilon, 1 + \varepsilon) \hat{A}_{\text{int}} \right) \right]\label{L_clip}
\end{equation} 


To further enhance the stability of the LLM output dynamic weights, we propose a semantic anchoring module that utilizes Retrieval-Augmented Generation (RAG \cite{lewis2020retrieval}) for stability optimization.
The proposed module utilizes a semantic reference corpus. This corpus is compiled from authoritative sources such as national standards, local regulations, industry guidelines, and technical specifications \cite{cai2024driving,yuan2024rag}.
Using a pretrained embedding model (text-embedding-ada-002), we embed both the driving conditions query and reference corpus into high-dimensional vector representations.
Semantic relevance between the query and corpus vectors is subsequently assessed using a FAISS-based similarity search \cite{johnson2019billion}.
A Top-3 selection strategy is employed to retrieve the most semantically aligned textual fragments.
The retrieved passages are aggregated into a semantic context set, which is then provided as auxiliary input to the LLM. 


This design leverages the LLM’s strength in context-aware weighting while mitigating the potential adverse effects of hallucinated outputs on Q-value evaluation, thereby suppressing policy misguidance caused by hallucinations.

\subsection{Semantic Cache Module}
The inference latency of LLMs inherently limits the control frequency of existing LLM-Dominated RL methods.
To address this problem, we propose a cooperative asynchronous training framework that coordinates low-frequency semantic planning by the LLM with the high-frequency control execution handled by the RL.
However, this decoupling introduces a new challenge: the LLM can fail to deliver timely semantic outputs since unexpected system delays or interruptions. 

To mitigate this, SCM compensates for such missing outputs by retrieving pertinent historical semantic representations.
The SCM maintains a dedicated memory bank.
This bank stores structured representations of past driving conditions, which encompass both scenario-level and object-level information.
Crucially, it also includes the corresponding multi-critic weight vectors that are generated in previously similar conditions.
When the LLM fails to return a valid semantic signal within a predefined time window, the SCM performs a rapid nearest-neighbor search over the memory bank using semantic embedding vectors.
The module then identifies the historical entry most semantically aligned with the current driving context. It extracts the associated weight vector from this entry to serve as a temporary guidance signal for the LLM.


\section{Experiments and Results}
\subsection{Experiment Setting}
\subsubsection{Driving Conditions}
All experiments are conducted in Town 2 of the CARLA simulator \cite{dosovitskiy2017carla}.
Driving conditions include conventional conditions, such as overtaking and merging, as well as safety-critical conditions, such as trilemma and occluded pedestrian.
To further assess the performance of the AD system, the traffic flow densities across three levels are established: low, medium, and high.

\subsubsection{Evaluation Metrics}
We evaluate the driving policy using quantitative metrics for safety, efficiency, and comfort.

\textbf{Safety} is evaluated using two primary metrics: \emph{Success Rate} (SR)—the percentage of episodes completed without major violations and \emph{Collision Rate} (CR)—the proportion of episodes involving collisions. 

\textbf{Efficiency} is evaluated by \emph{Average Speed} (AS), \emph{Total Distance} (TD) and \emph{Time Steps} (TS), reflecting travel speed, distance covered, and task completion time, respectively.

\textbf{Comfort} is evaluated via \emph{Speed Variance} (SV) and \emph{Acceleration Variance} (AV), which reflects the smoothness and stability of driving behavior. 

\subsubsection{Baselines}

We systematically compare the proposed method with the following methods:
\begin{itemize}[leftmargin=1.5em]
    \item \textbf{Vanilla PPO} \cite{schulman2017proximal}: Directly train the policy using PPO in tasks as a basic RL baseline.
    \item \textbf{E2ECLA} \cite{anzalone2022end}: Combine curriculum learning and RL, which learns end-to-end AD policies in CARLA by gradually increasing task difficulty without prior knowledge.
    \item \textbf{AutoReward} \cite{han2024autoreward}: An RL method that iteratively refines LLM-generated rewards post-training.
    \item \textbf{VLM-RL} \cite{huang2024vlm}: A method that integrates pre-trained VLM with RL. It generates semantic rewards through language objective comparison, replacing manually designed reward function.
\end{itemize}

\subsection{Main Results}

As illustrated in Table ~\ref{tab:main_results}, SR and CR of various methods are compared in conventional conditions under different traffic densities.
Results indicate that HCRMP matches or outperforms other baselines in different driving conditions, particularly in medium- and high-density conditions, where it achieves an average SR of 89.5\%.
This notable performance can be attributed to the incorporation of CSA, which dynamically enhances the emphasis on safety. 
By optimizing the multi-critic coordination strategy, the agent is able to adopt safer actions, significantly reducing the collision risk.


Additionally, HCRMP shows a relatively minor advantage over the baselines in low-density driving conditions, which is due to limited vehicle interactions. It reduces the difficulty of the driving task and restrict the full exploitation of the CSA's dynamic adjustment capabilities.

\begin{table*}[htbp] 
\centering
\caption{Performance Comparison in Conventional Conditions under Different Traffic Densities} 
\fontsize{7.5}{9}\selectfont
\label{tab:main_results} 
\begin{tabular}{l l l c c c c c c} 
    \toprule
    \multirow{2}{*}{Methods}  & \multirow{2}{*}{Category}& \multirow{2}{*}{Condition} & \multicolumn{2}{c}{Low Density} & \multicolumn{2}{c}{Medium Density} & \multicolumn{2}{c}{High Density} \\
    \cmidrule(lr){4-5} \cmidrule(lr){6-7} \cmidrule(lr){8-9} 
    && & SR (\%) & CR (\%) & SR (\%) & CR (\%) & SR (\%) & CR (\%) \\
    \midrule
    Vanilla PPO & RL&Overtaking & 80.0 & 20.0 & 69.0 & 31.0 & 62.0 & 38.0 \\
              &  & Merging    & 89.0 & 11.0 & 75.0 & 25.0 & 68.0 & 32.0 \\
     \midrule
    E2ECLA &RL& Overtaking & 60.0 & 40.0 & 56.0 & 44.0 & 38.0 & 62.0 \\
           && Merging    & 56.0 & 44.0 & 54.0 & 46.0 & 42.0 & 58.0 \\
    \midrule
    AutoReward (iter=0) &LLM-Dominated RL& Overtaking & 73.0 & 27.0 & 60.0 & 40.0 & 48.0 & 52.0 \\
                     & & Merging    & 82.0 & 18.0 & 71.0 & 29.0 & 51.0 & 49.0 \\
    \midrule
    AutoReward (iter=5) &LLM-Dominated RL& Overtaking & 85.0 & 15.0 & 76.0 & 24.0 & 70.0 & 30.0 \\
                     & & Merging    & 94.0 & 6.0 & 85.0 & 15.0 & 71.0 & 29.0 \\
    \midrule
    VLM-RL &LLM-Dominated RL& Overtaking & 54.0 & 45.0 & 52.0 & 48.0 & 50.0 & 50.0 \\
           && Merging    & 56.0 & 44.0 & 53.0 & 47.0 & 50.0 & 50.0 \\
    \midrule
    HCRMP  &LLM-Hinted RL& Overtaking & \textbf{99.0} & \textbf{1.0} & \textbf{93.0} & \textbf{7.0} & \textbf{87.0} & \textbf{13.0} \\
           && Merging    & 97.0 & 3.0 & 92.0 & 8.0 & 86.0 & 14.0 \\
    \bottomrule
\end{tabular}
\end{table*}

We further conduct a systematic evaluation of HCRMP and three baselines-E2ECLA, VLM-RL, and Autoreward—under safety-critical scenarios, with the latter two being representative LLM-Dominated RL methods.
The corresponding results are presented in Table~\ref{tab:critical_results_updated}.

\begin{table*}[h!] 
\centering
\caption{Performance Comparison in Safety-Critical Driving Conditions} 
\label{tab:critical_results_updated} 
\setlength{\tabcolsep}{3pt} 
\small
\begin{tabular}{l l l c c c c c c c} 
    \toprule
    \multirow{2}{*}{Methods}& \multirow{2}{*}{Condition} & \multirow{2}{*}{\shortstack[c]{Traffic\\Density}} & SR & CR & AS & TD & TS & SV & AV \\
    & & & (\%) & (\%) & (m/s) & (m) & (s) & (m/s) & (m/s$^2$) \\
    \midrule
    \multirow{6}{*}{E2ECLA}
     & \multirow{3}{*}{\shortstack[c]{Occluded\\Pedestrian}} & Low    & 36.0          & 64.0          & 8.22          & 36.99          & 47.04          &  4.29  &  \textbf{1.49}  \\
     &                                                       & Medium & 37.0          & 63.0          & 7.19          & 31.53          & 39.72          &  3.02  &  2.62           \\
     &                                                       & High   & 31.0          & 69.0          & 7.87          & 33.86          & 37.04          &  3.25  &  \textbf{1.52}  \\
    \cmidrule{2-10} 
     & \multirow{3}{*}{Trilemma}                             & Low    & 36.0          & 64.0          & 6.99          & 52.82          & 48.38          &  3.01  &  2.89           \\
     &                                                       & Medium & 34.0          & 66.0          & 6.99          & 40.64          & \textbf{50.28} &  \textbf{2.18}  &  2.70           \\
     &                                                       & High   & 38.0          & 62.0          & 8.68          & 48.96          & 46.44          &  4.59  &  2.89           \\
    \midrule
    \multirow{6}{*}{Autoreward(iter=5)}
     & \multirow{3}{*}{\shortstack[c]{Occluded\\Pedestrian}} & Low    & 70.0          & 30.0          & 6.30          & 28.07 & 35.05          &  \textbf{1.40}           &  2.96           \\
     &                                                       & Medium & 31.0          & 69.0          & 7.79         & 27.42 & \textbf{55.78}          & \textbf{ 1.34}           &  3.16           \\
     &                                                       & High   & 24.0          & 76.0          & 7.37          & 31.26& \textbf{53.65}          & \textbf{ 2.15}           &  3.21
     \\
    \cmidrule{2-10} 
     & \multirow{3}{*}{Trilemma}                             & Low    & 60.0          & 40.0          & 5.60          & 24.91 & 34.3          &  \textbf{2.45 }          &  2.68           \\
     &                                                       & Medium & 58.0          & 42.0          & 6.16          & 31.22 & 33.9         &  3.05           &  3.33           \\
     &                                                       & High   & 32.0          & 68.0          & 7.05          & 25.84 & \textbf{55.27}          &  \textbf{1.90}           &  2.95           \\
    \midrule
    \multirow{6}{*}{VLM-RL}
     & \multirow{3}{*}{\shortstack[c]{Occluded\\Pedestrian}} & Low    & 65.0          & 35.0          & 5.02          & \textbf{139.20}& 45.04     &  8.43           &  1.72           \\
     &                                                       & Medium & 53.0          & 47.0          & 6.34          & \textbf{170.45}& 42.30          &  7.69           &  1.78           \\
     &                                                       & High   & 41.0          & 59.0          & 5.98          & \textbf{172.87}& 40.23          &  7.06           &  1.61           \\
    \cmidrule{2-10} 
     & \multirow{3}{*}{Trilemma}                             & Low    & 67.0          & 33.0          & 8.89          & \textbf{283.13}& 50.01          &  9.87           &  1.75           \\
     &                                                       & Medium & 58.0          & 42.0          & 7.76          & \textbf{278.82}& 48.13          &  7.75           &  1.69           \\
     &                                                       & High   & 54.0          & 46.0          & 7.79          & \textbf{176.06}& 46.98          &  7.72           &  2.70           \\
    \midrule
    \multirow{6}{*}{HCRMP}
     & \multirow{3}{*}{\shortstack[c]{Occluded\\Pedestrian}} & Low    & \textbf{73.0} & \textbf{27.0} & \textbf{9.98} & 86.06          & \textbf{50.07} &  10.04          &  1.69           \\
     &                                                       & Medium & \textbf{67.0} & \textbf{33.0} & \textbf{9.96} & 82.09          & 51.17 &  9.97           &  \textbf{1.74}  \\
     &                                                       & High   & \textbf{61.0} & \textbf{39.0} & \textbf{8.97} & 77.95          & 48.59 &  9.69           &  1.72           \\
    \cmidrule{2-10} 
     & \multirow{3}{*}{Trilemma}                             & Low    & \textbf{75.0} & \textbf{28.0} & \textbf{10.24}& 88.20          & \textbf{51.58} &  9.57           &  \textbf{1.44}  \\
     &                                                       & Medium & \textbf{69.0} & \textbf{31.0} & \textbf{10.08}& 79.91          & 49.72          &  10.14          &  \textbf{1.23}  \\
     &                                                       & High   & \textbf{64.0} & \textbf{36.0} & \textbf{9.94} & 77.34          & 47.13 &  9.96           &  \textbf{1.72}  \\
    \bottomrule
\end{tabular}
\end{table*}



E2ECLA exhibits an AV exceeding 2.5 m/s² under certain conditions, which can cause discomfort to the passenger \cite{de2023standards}. 
This phenomenon can be primarily attributed to E2ECLA's failure to adequately consider vehicle acceleration. Consequently, the system prioritizes rapid maneuvers over comfort. 
AutoReward utilizes LLM to construct the reward function based on the analysis of the scenario.
Similarly, unreasonable values of acceleration changes occur due to the LLM's emphasis on maneuverability at the expense of comfort considerations in safety-critical scenarios.
VLM-RL, leveraging efficient navigation approaches, achieves a notably high TD; however, its average CR in safety-critical conditions reaches 43.7\%, indicating a significant safety concern. 

The proposed HCRMP demonstrates a well-balanced performance across all metrics, particularly in high-density conditions, where it achieves a SR of 62.5\%, a CR of 37.5\%, and an AV of 1.72 m/s². This comprehensive performance is mainly due to the integration of ASR and CSA. The former pursues efficiency and comfort in low-density traffic flows, while prioritizing safety in medium- and high-density traffic conditions. The latter enhances the system's ability to understand complex environments by extending the state space.


HCRMP prioritizes ensuring safety in immediate driving tasks.
This focus, however, means its global adaptation to the map's inherent static path features may be less developed, which consequently results in a lower traveled distance compared to VLM-RL. 
Through the synergistic effect of CSA and ASR, HCRMP achieves an effective balance of safety, efficiency, and comfort in safety-critical conditions, with particularly outstanding performance in medium- and high-density conditions.

\subsection{Ablation Study}
Table ~\ref{tab:ablation_results} presents the results of the ablation study in the medium-density trilemma, evaluating the performance of the HCRMP with the removal of different modules: HCRMP without ASR, HCRMP without CSA, and HCRMP with ASR.
The SR of HCRMP without ASR drops to a mere 40.0\%, indicating that the absence of ASR significantly diminishes the system's ability to comprehend complex environments, resulting in information deficits that increase collision risks.
In contrast, HCRMP with ASR, achieves an SR of 54.0\%, underscoring the critical role of ASR in enhancing situational awareness. 
By leveraging the LLM to interpret the current driving conditions, ASR expands the state space, strengthening the system's awareness of surrounding driving risks and thereby improving the safety of its decision-making tasks.

Furthermore, HCRMP without CSA exhibits an SR of 48.0\%, which is even lower than that of HCRMP with ASR. Specifically, CSA enhances contextual stability through a knowledge base by dynamically constraining the priority weights within the multi-critic
framework, effectively mitigating excessive fluctuations during the training process. Without CSA, the multi-critic
system experiences pronounced instability in weight adjustments, hindering convergence toward an optimal policy. This instability directly undermines the system's decision-making consistency and safety in the medium-density trilemma, consequently leading to a substantial decline in SR.

\begin{table}[htbp] 
\centering
\caption{Ablation Study Results} 
\label{tab:ablation_results} 
\tiny
\begin{tabular}{l c c c c c c c} 
    \toprule
    Model & SR(\%) & CR(\%) & AS(m/s) & TD(m) & TS(s) & SV(m/s) & AV(m/s$^2$) \\
    \midrule
    HCRMP w/o ASR & 40 & 60    & \textbf{7.52}    &   \textbf{44.34}    & 47.74      & 6.94     & 2.95     \\ 
    HCRMP w/o CSA & 48 & 52     & 6.01      & 29.82      & \textbf{54.44}      & \textbf{2.89}     & 2.63     \\  
    HCRMP w/ ASR  & \textbf{54}     & \textbf{46}     & 7.27      &   29.04    & 34.64      &  5.89    &    \textbf{2.26}  \\ 
    \bottomrule
\end{tabular}
\end{table}

\begin{figure}[H]
    \centering
    \includegraphics[width=0.9\linewidth]{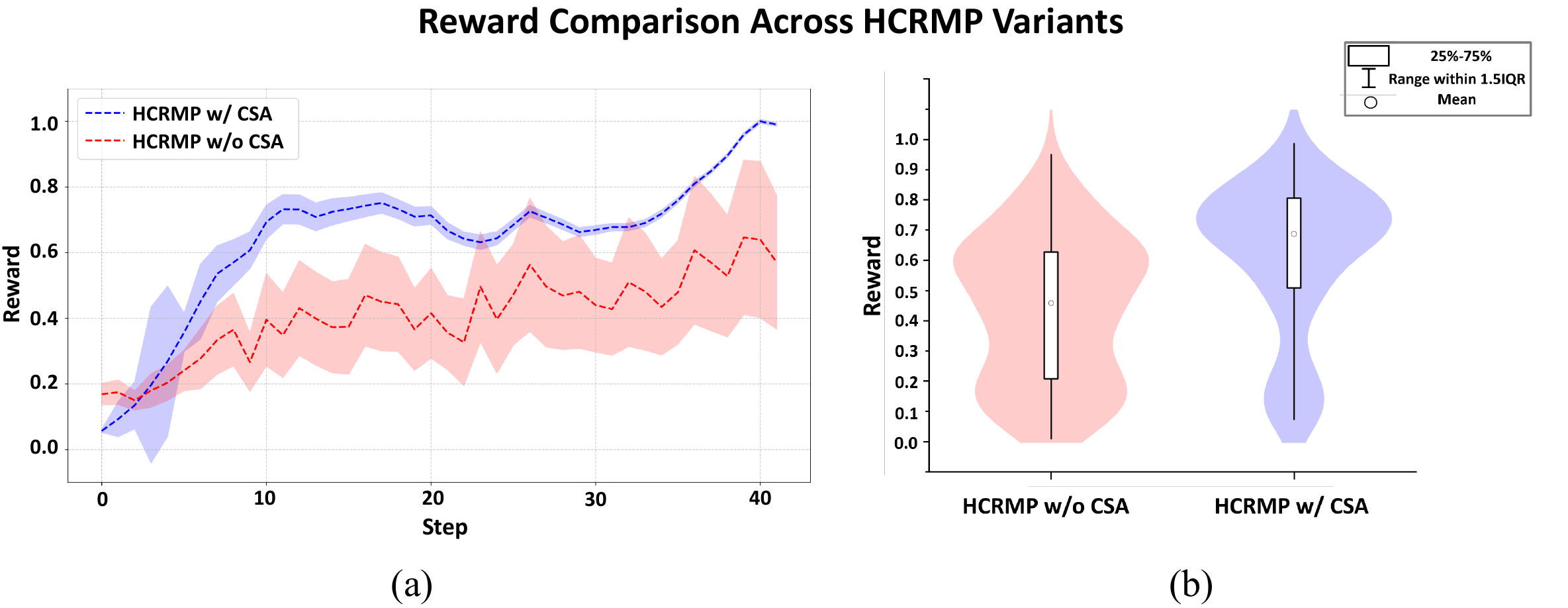}
    \caption{HCRMP variants rewards: dynamic trends and statistical distributions.
    Figure (a) visualizes that the dynamic reward curve for HCRMP without CSA exhibits significant fluctuations, indicative of performance instability, while the curve for HCRMP with CSA shows smaller fluctuations and more stable performance.
    Figure (b) illustrates that the overall reward distribution for the variant without CSA is wider and more pronounced in lower reward regions, whereas the CSA variant's rewards are more concentrated in higher value ranges with a more prominent peak.}
    \label{fig:reward_curve}
\end{figure}

Figure~\ref{fig:reward_curve} illustrates the performance differences between HCRMP variants through dynamic reward trends.
The reward curve for HCRMP without CSA exhibits pronounced fluctuations, indicating that, in the absence of CSA constraints, the system struggles to stabilize reward values during evaluating, reflecting inherent instability in policy optimization. 
In contrast, the HCRMP variant equipped with CSA demonstrates a smoother upward trend in its reward curve.




\section{Conclusion}
Current LLM-Dominated RL approaches for AD motion planning heavily depend on LLM outputs, making them vulnerable to hallucinations that can compromise policy reliability and lead to unsafe behavior.
We propose a LLM-Hinted RL motion planning paradigm and the corresponding HCRMP framework, aiming to preserve the relative independence between LLM and RL. 
The framework mitigates the impact of LLM hallucinations, while still preserving the strengths of LLM in semantic understanding and high-level decision-making while ensuring RL maintains its fundamental self-optimization capabilities.
The HCRMP architecture comprises three key components.
First, the \ding{172} \textbf{Augmented Semantic Representation Module} refines the state space via semantic guidance.
Second, the \ding{173} \textbf{Contextual Stability Anchor Module} enhances the reliability of LLM-provided multi-critic weight hints through retrieval-augmented semantic anchoring based on information from the structured knowledge base.
Finally, the \ding{174} \textbf{Semantic Cache Module} primarily improves training efficiency by asynchronous decoupling low-frequency LLM reasoning from high-frequency RL execution.
To handle delayed LLM outputs, it employs a historical semantic cache matching strategy as a fallback.
Extensive experiments in CARLA validate HCRMP's strong overall driving performance.
HCRMP has a high task success rate of 80.3\% under diverse driving conditions with different traffic densities. 
Especially, it achieves a 11.4\% reduction in collision rate across a range of complex driving conditions.
HCRMP provides a promising framework for RL motion planning for AD with integrated LLM.

\bibliographystyle{unsrt}
\bibliography{HCRMP_Arxiv.bbl}

\end{document}